\documentclass[english,reqno]{article}
\usepackage{mathptmx}
\usepackage[T1]{fontenc}
\usepackage[latin9]{inputenc}
\usepackage{geometry}
\usepackage{wrapfig}
\usepackage{amsmath}
\usepackage{amsthm}
\usepackage{graphicx}
\usepackage[authoryear]{natbib}

\makeatletter

\newcommand{\lyxdot}{.}

\theoremstyle{plain}
\newtheorem{thm}{\protect\theoremname}
  \theoremstyle{definition}
  \newtheorem{problem}[thm]{\protect\problemname}

\renewcommand{\vec}[1]{{\bf #1}}

\makeatother

\usepackage{babel}
  \providecommand{\problemname}{Problem}
\providecommand{\theoremname}{Theorem}

\begin{document}

\title{A correlation game for unsupervised learning yields computational
interpretations of Hebbian excitation, anti-Hebbian inhibition, and
synapse elimination }

\author{Sebastian Seung and Jonathan Zung\\
Neuroscience Institute and Computer Science Dept.\\
Princeton University\\
Princeton, NJ 08544}
\maketitle
\begin{abstract}
Much has been learned about plasticity of biological synapses from
empirical studies. Hebbian plasticity is driven by correlated activity
of presynaptic and postsynaptic neurons. Synapses that converge onto
the same neuron often behave as if they compete for a fixed resource;
some survive the competition while others are eliminated. To provide
computational interpretations of these aspects of synaptic plasticity,
we formulate unsupervised learning as a zero-sum game between Hebbian
excitation and anti-Hebbian inhibition in a neural network model.
The game formalizes the intuition that Hebbian excitation tries to
maximize correlations of neurons with their inputs, while anti-Hebbian
inhibition tries to decorrelate neurons from each other. We further
include a model of synaptic competition, which enables a neuron to
eliminate all connections except those from its most strongly correlated
inputs. Through empirical studies, we show that this facilitates the
learning of sensory features that resemble parts of objects. 
\end{abstract}
Much has been learned about plasticity of biological synapses from
empirical studies. One line of research has explored Hebbian plasticity,
which is triggered by correlated presynaptic and postsynaptic activity.
Many kinds of excitatory synapses are strengthened by correlated activity,
and are said to be Hebbian \citep{bi2001synaptic}. Strengthening
of inhibitory synapses by correlated activity has also been observed
\citep{gaiarsa2002long}. The phenomenon is sometimes said to be anti-Hebbian,
because strengthening an inhibitory synapse is like making a negative
number more negative. Another line of research has shown that synapses
converging onto the same neuron often behave as if they compete for
shares of a fixed ``pie''; some survive the competition while others
are eliminated \citep{lichtman2000synapse}. According to this viewpoint,
Hebbian plasticity does not increase or decrease the pie of synaptic
resources; it only allocates resources across convergent synapses
\citep{miller1996synaptic}

Theoretical neuroscientists have proposed a number of computational
functions for Hebbian excitation, anti-Hebbian inhibition, and synaptic
competition/elimination. Hebbian excitation has long been invoked
as a mechanism for learning features from sensory input \citep{von1973self}.
Lateral inhibition has been used to sparsen neural activity, thereby
facilitating Hebbian feature learning \citep{von1973self,fukushima1980neocognitron,rumelhart1985feature,kohonen1990self}.
Endowing the lateral inhibition with anti-Hebbian plasticity can give
more robust control over sparseness of activity. \citet{foldiak1990forming}
demonstrated this with numerical experiments, but did not provide
an interpretation in terms of an optimization principle. Less relevant
here are anti-Hebbian models without reference to sparse activity
in linear \citep{foldiak1989adaptive,rubner1989self,rubner1990development}
and nonlinear \citep{carlson1990anti,girolami1997extended} networks.
Also less relevant is the application of anti-Hebbian plasticity to
feedforward rather than lateral connections \citep{hyvarinen1998independent}.

\citet{leen1991dynamics} performed a stability analysis for a linear
network with Hebbian feedforward connections and anti-Hebbian lateral
connections. \citet{plumbley1993efficient} derived a linear network
with anti-Hebbian lateral inhibition (but no plasticity of feedforward
connections) from the principle of information maximization with a
power constraint. \citet{pehlevan2015hebbian} showed that a linear
network with Hebbian feedforward connections and anti-Hebbian lateral
inhibition can be interpreted as online gradient optimization of a
``similarity matching'' cost function. \citet{pehlevan2014hebbian}
and \citet{hu2014hebbian} went on to extend the similarity matching
principle to derive nonlinear neural networks for unsupervised learning.
Synaptic competition and elimination have been studied in models of
cortical development, and have been shown to play an important role
in the emergence of feature selectivity \citep{miller1996synaptic}. 

The subject of the present work is a mathematical formalism that provides
computational interpretations of Hebbian excitation, anti-Hebbian
inhibition, and synaptic competition/elimination in nonlinear neural
networks. We start by formulating unsupervised learning as the maximization
of output-input correlations subject to upper bound constraints on
output-output correlations. We motivate our formulation by describing
its relation to previous theoretical frameworks for unsupervised learning,
such as maximization \citep{linsker1988self,atick1990towards,plumbley1993efficient,bell1995information}
or minimization of mutual information \citep{hyvarinen2000independent},
(2) projection onto a subspace that maximizes a moment-based statistic
such as variance \citep{oja1982simplified,linsker1988self} or kurtosis
\citep{huber1985projection}, and the (3) similarity matching principle
\citep{pehlevan2015hebbian}.

To solve our constrained maximization problem, we introduce Lagrange
multipliers. This Lagrangian dual formulation of unsupervised learning
can in turn be solved by a nonlinear neural network with Hebbian excitation
and anti-Hebbian inhibition. The network is very similar to the original
model of \citet{foldiak1990forming}, differing mainly by its use
of rectification rather than sigmoidal nonlinearity. (The latter can
also be handled by our formalism, as shown in Appendix \ref{sec:FoldiakOriginal}.)
Lagrange multipliers were also used to study anti-Hebbian plasticity
by \citet{plumbley1993efficient}, but only for linear networks.

Effectively, excitation and inhibition behave like players in a game,
and the inhibitory connections can be interpreted as Lagrange multipliers.
The game is zero-sum, in that excitation tries to maximize a payoff
function and inhibition tries to minimize exactly the same payoff
function. Roughly speaking, however, one could say that excitation
aims to maximize the correlation of each output neuron with its inputs,
while inhibition aims to decorrelate the output neurons from each
other. Our term ``correlation game'' is derived from this intuitive
picture. 

Within our mathematical formalism, we also consider a dynamics of
synaptic competition and elimination that is drawn from models of
cortical development \citep{miller1994role}. Competition between
the excitatory synapses convergent on a single neuron is capable of
driving the strengths of some synapses to zero. In numerical experiments
with the MNIST dataset, we show that synapse elimination has the computational
function of facilitating the learning of features that resemble ``parts''
of objects. Theoretical analysis shows that the surviving synapses
converging onto an output neuron come from its most strongly correlated
inputs; synapses from weakly correlated inputs are eliminated. 

Our correlation game is closely related to the similarity matching
principle of \citet{pehlevan2014hebbian} and \citet{pehlevan2015hebbian}.
The major novelty is the introduction of decorrelation as a constraint
for the optimization. Paralleling our work, \citet{1703.07914} have
shown that the similarity matching principle leads to a game theoretic
formulation through Hubbard-Stratonovich duality. Again our novelty
is the use of decorrelation as a constraint, which leads to our correlation
game through Lagrangian duality.

Our model of synaptic competition and elimination was borrowed with
slight modification from the literature on modeling neural development
\citep{miller1994role}. It can be viewed as a more biologically plausible
alternative to previous unsupervised learning algorithms that sparsen
features. For example, \citet{hoyer2004non} is similar to ours because
it can be interpreted as independently sparsening each set of convergent
synapses, rather than applying a global L1 regularizer to all synapses.

\section{Primal formulation}

Our neural networks will learn to transform a sequence of input vectors
$\vec{u}(1),\hdots,\vec{u}(T)$ into a sequence of output vectors
$\vec{x}(1),\hdots,\vec{x}(T)$. Both input and output will be assumed
nonnegative.

Define the input matrix $U=\left[\vec{u}(1),\ldots,\vec{u}\left(T\right)\right]$
as the matrix containing input vectors $\vec{u}(t)$ as its columns.
The element $U_{at}$ is the $a$th component of $\vec{u}(t)$. Similarly,
define the output matrix $X=\left[\vec{x}\left(1\right),\ldots,\vec{x}\left(T\right)\right]$
as containing output vectors $\vec{x}(t)$ as its columns. Both input
and output will be assumed nonnegative. We define the output-input
correlation matrix as
\[
\frac{XU^{\top}}{T}=\frac{1}{T}\sum_{t=1}^{T}\vec{x}\left(t\right)\vec{u}\left(t\right)^{\top}
\]
Its $ia$ element is the time average of $x_{i}u_{a}$, or $\langle x_{i}u_{a}\rangle$.
Similarly, we define the output-output correlation matrix
\[
\frac{XX^{\top}}{T}=\frac{1}{T}\sum_{t=1}^{T}\vec{x}\left(t\right)\vec{x}\left(t\right)^{\top}
\]
Its $ij$ element is the time average of $x_{i}x_{j}$, or $\langle x_{i}x_{j}\rangle$.
Note that we use ``correlation matrix'' to mean second moment matrix
rather than covariance matrix. In other words, our correlation matrix
does not involve subtraction of mean values. We believe that this
is natural for sparse nonnegative variables, but covariance matrices
may be substituted in other settings (see Appendix \ref{sec:FoldiakOriginal}).
\begin{problem}
[Primal formulation] \label{prob:Primal}We define the goal of unsupervised
learning as the constrained optimization
\end{problem}
\begin{equation}
\max_{X\geq0}\Phi^{\ast}\left(\frac{XU^{\top}}{T}\right)\text{ such that }\frac{XX^{\top}}{T}\leq D\label{eq:ConstrainedOptimization}
\end{equation}
where $D$ is a fixed matrix and $\Phi^{\ast}$ is a scalar-valued
function of a matrix argument. We will assume that $\Phi^{\ast}(C)$
is monotone nondecreasing as a function of every element of $C$,
which allows us to interpret the objective of Eq. (\ref{eq:ConstrainedOptimization})
as maximization of correlations between inputs and outputs. Later
on, it will also be convenient to assume that $\Phi^{\ast}$ is convex. 

The inequality constraint in Eq. (\ref{eq:ConstrainedOptimization})
applies to all elements of the matrices $XX^{\top}/T$ and $D$. Its
role is to prevent two kinds of degenerate optima. The diagonal elements
of the constraint, $\langle x_{i}^{2}\rangle\leq D_{ii}$, limit the
power in the outputs, as in \citet{plumbley1993efficient}. This prevents
the trivial solution of maximizing output-input correlations by scaling
up $x_{i}$. The off-diagonal elements, $\langle x_{i}x_{j}\rangle\leq D_{ij}$,
insure that the outputs remain distinct from each other. Suppose for
example that all diagonal elements of $D$ are $q^{2}$ and all off-diagonal
elements of $D$ are $p^{2},$
\begin{equation}
D_{ij}=\begin{cases}
p^{2}, & i\neq j,\\
q^{2}, & i=j
\end{cases}\label{eq:DesiredCorrelations}
\end{equation}
If a pair of outputs saturates the inequality constraints, then their
cosine similarity is
\begin{equation}
\frac{\left\langle x_{i}x_{j}\right\rangle }{\sqrt{\langle x_{i}^{2}\rangle}\sqrt{\langle x_{j}^{2}\rangle}}=\frac{p^{2}}{q^{2}}\label{eq:CosineSimilarity}
\end{equation}
If the outputs are nonnegative, the cosine similarity lies between
0 and 1. A cosine similarity of $p/q=1$ means that every output is
exactly the same. A cosine similarity of $p/q=0$ means a ``winner-take-all''
representation in which only one output is active for any given stimulus.
We will generally work in the intermediate regime where $p^{2}/q^{2}$
is positive but much less than one. In this regime, we can say that
the constraint in Eq. (\ref{eq:ConstrainedOptimization}) requires
that the outputs be ``decorrelated'' or ``desimilarized.'' The
outputs end up sparse but distributed, as will be seen empirically
later on. To summarize the above, we can say that Eq. (\ref{eq:ConstrainedOptimization})
maximizes output-input correlations while decorrelating the outputs.

Our Problem \ref{prob:Primal} may look unfamiliar, but in fact it
is similar to other formulations of unsupervised learning that are
already well-accepted: (1) maximization \citep{linsker1988self,atick1990towards,bell1995information}
or minimization \citep{hyvarinen2000independent} of mutual information,
(2) projection onto a subspace that maximizes a moment-based statistic
such as variance or kurtosis \citep{huber1985projection}, and (3)
the similarity matching principle \citep{pehlevan2015hebbian}. These
connections are discussed in Appendix \ref{sec:OtherObjectives}.
One might ask whether a wider class of objective functions admits
a similar analysis to the one given in this paper. An example of the
flexibility of our approach is given in Appendix \ref{sec:FoldiakOriginal}. 

\section{Lagrangian dual and the correlation game}

While Problem \ref{prob:Primal} has connections with traditional
unsupervised learning objectives, a dual formulation makes its connections
to neural networks manifest. We make use of two duality transforms
to introduce auxiliary variables $W_{ia}$ and $L_{ij}$, which will
respectively turn out to be feedforward and lateral synaptic weight
matrices of a neural network.

First, we rewrite $\Phi^{\ast}$ in terms of its Legendre-Fenchel
transform $\Phi$,

\begin{align}
\Phi^{\ast}(C) & =\max_{W}\left\{ \sum_{ia}W_{ia}C_{ia}-\Phi\left(W\right)\right\} \label{eq:ConvexConjugate}\\
 & =\max_{W\geq0}\left\{ \sum_{ia}W_{ia}C_{ia}-\Phi\left(W\right)\right\} 
\end{align}
The first equality holds since $\Phi^{\ast}$ is convex, and the second
equality holds since $\Phi^{\ast}$ is monotonic as a function of
each element of $C$. We will often take $\Phi(W)=\infty$ outside
a convex set $B$. This effectively restricts $W$ and lets us replace
$\max_{W\geq0}$ with $\max_{W\in B}$. 

Second, we introduce Lagrange multipliers $L_{ij}$ to solve the constrained
optimization of Eq. (\ref{eq:ConstrainedOptimization}). The optimization
is equivalent to
\begin{equation}
\max_{X\geq0}\min_{L\geq0}\left\{ \Phi^{\ast}\left(\frac{XU^{\top}}{T}\right)-\frac{1}{2}\sum_{ij}L_{ij}\left(\frac{1}{T}\sum_{t}X_{it}X_{jt}-D_{ij}\right)\right\} \label{eq:Primal}
\end{equation}
(The outer maximum must choose $X$ such that $XX^{\top}/T\leq D$
because otherwise the minimum with respect to $L$ is $-\infty$.)
Switching the order of maximum and minimum yields the following unconstrained
optimization. 
\begin{problem}
[Dual formulation] \label{prob:Dual}An alternative to Problem \ref{prob:Primal}
is to define unsupervised learning as
\end{problem}
\begin{equation}
\min_{L\geq0}\max_{X\geq0}\left\{ \Phi^{\ast}\left(\frac{XU^{\top}}{T}\right)-\frac{1}{2}\sum_{ij}L_{ij}\left(\frac{1}{T}\sum_{t}X_{it}X_{jt}-D_{ij}\right)\right\} \label{eq:Dual}
\end{equation}
By the minimax inequality, this is an upper bound for Eq. (\ref{eq:Primal}).
A maximization with respect to $W$ is implicit in the definition
(\ref{eq:ConvexConjugate}) of the convex conjugate $\Phi^{\ast}$.
We can switch the order of $W$ and $X$ maximizations to obtain the
following equivalent problem.
\begin{problem}
[Correlation game] Define the payoff function
\end{problem}
\begin{equation}
R(W,L)=\max_{X\geq0}\left\{ \frac{1}{T}\sum_{t}\left[\sum_{ia}W_{ia}X_{it}U_{at}-\Phi\left(W\right)-\frac{1}{2}\sum_{ij}L_{ij}\left(X_{it}X_{jt}-D_{ij}\right)\right]\right\} \label{eq:Payoff}
\end{equation}
Then Eq. (\ref{eq:Dual}) is equivalent to 
\begin{equation}
\min_{L\geq0}\max_{W\in B}R(W,L)\label{eq:CorrelationGame}
\end{equation}
which can be interpreted as a zero-sum game played between excitation
and inhibition. The goal of excitation is to maximize the payoff function
(\ref{eq:Payoff}), and the goal of inhibition is to minimize the
same payoff function. Roughly speaking, however, one could say that
excitation aims to maximize output-input correlations while inhibition
aims to decorrelate the outputs. This is our rationale for referring
to (\ref{eq:CorrelationGame}) as the ``correlation game.''

\section{Neural network algorithm}

\begin{wrapfigure}{O}{2in}%
\includegraphics[width=2in]{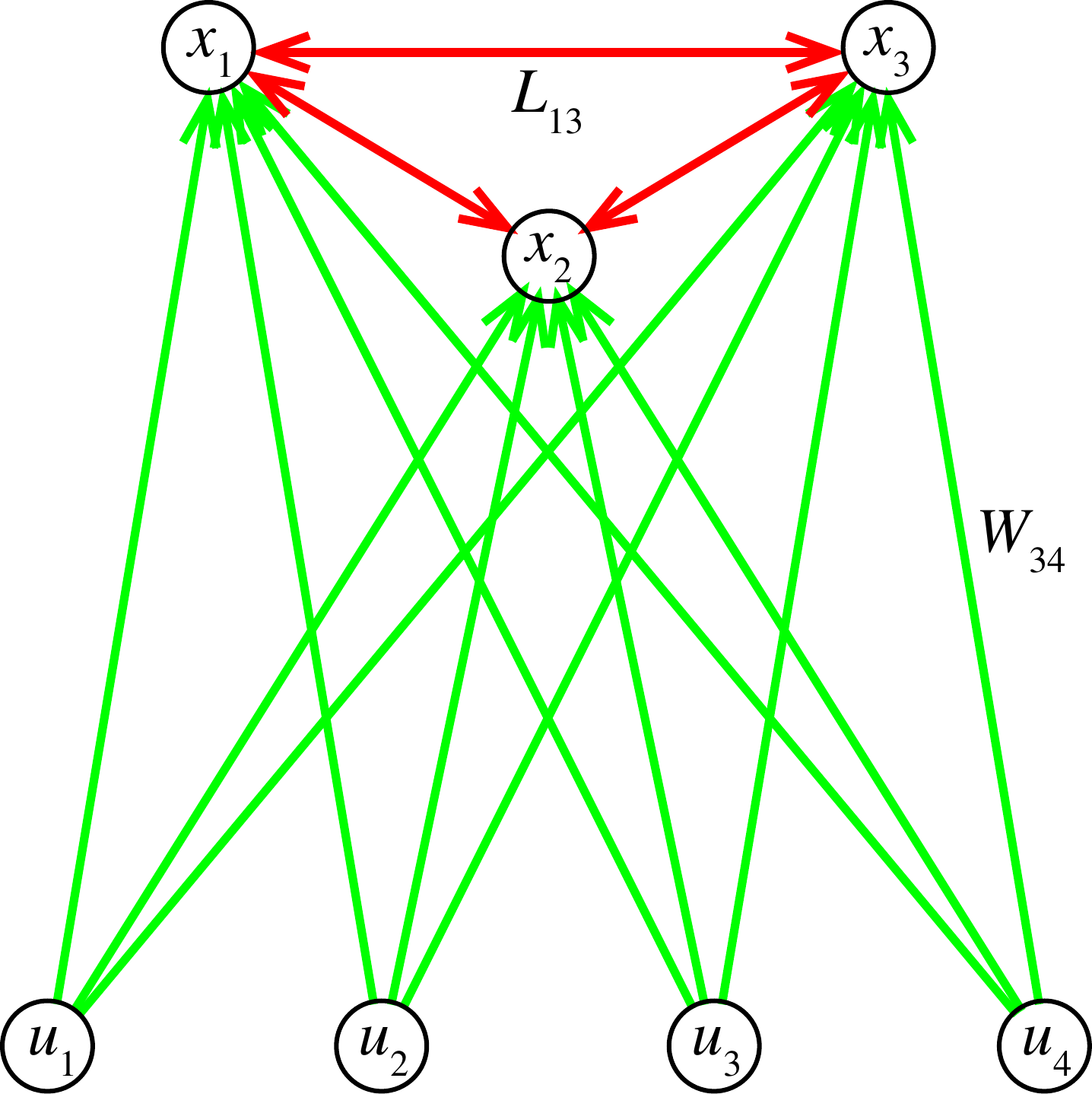}

\caption{Architecture of network with Hebbian feedforward excitation (green)
and anti-Hebbian lateral inhibition (red).\label{fig:Architecture}}
\end{wrapfigure}%
The optimizations in Eq. \ref{eq:CorrelationGame} could be performed
by many methods. Here we consider an iterative online method, which
is based on one input vector at each time step. First we describe
the optimization with respect to $X$ in Eq. (\ref{eq:Payoff}). The
objective function is nondecreasing under coordinate ascent with nonnegativity
constraints,
\begin{equation}
X_{it}:=\frac{1}{L_{ii}}\left[\sum_{a}W_{ia}U_{at}-\sum_{j,j\neq i}L_{ij}X_{jt}\right]^{+}\label{eq:CoordinateAscent}
\end{equation}
This can be interpreted as the dynamics of a neural network (Fig.
\ref{fig:Architecture}). The conjugate variables $W_{ia}$ in the
Legendre-Fenchel transform of Eq. (\ref{eq:ConvexConjugate}) are
now feedforward connections from the input to the output. The Lagrange
multipliers $L_{ij}$ are now lateral connections between the outputs.
The lateral connections are assumed to be symmetric ($L_{ij}=L_{ji})$,
which guarantees that the dynamics will converge to a local maximum
of the objective function if there is no runaway instability \citep{hahnloser2003permitted}.
If the diagonal elements of $L$ have a positive lower bound and
the off-diagonal elements are nonnegative, then $L$ is copositive
definite, $\vec{x}^{T}L\vec{x}\geq0$ for $\vec{x}\geq0$ with equality
only for $\vec{x}=0$. It follows that the dynamics of Eq. (\ref{eq:CoordinateAscent})
exhibits no runaway instability \citep{hahnloser2003permitted}. The
coordinate ascent dynamics of Eq. (\ref{eq:CoordinateAscent}) was
previously considered by \citet{pehlevan2014hebbian}, who added nonnegativity
constraints to the similarity matching principle. Coordinate ascent
is a particularly simple way of optimizing Eq. (\ref{eq:Payoff})
with respect to $X$; many other ``neural'' algorithms could be
used. Appendix \ref{sec:FoldiakOriginal} extends our formalism to
neural networks with sigmoidal rather than rectification nonlinearity.

For the other optimizations in Eq. (\ref{eq:CorrelationGame}), we
perform gradient ascent with respect to $W$ and gradient descent
with respect to $L$. Stochastic projected gradient ascent on $W$
is 
\begin{equation}
\Delta W_{ia}\propto X_{it}U_{at}-\frac{\partial\Phi}{\partial W_{ia}}\label{eq:StochasticGradientAscent}
\end{equation}
followed by projection of $W$ onto the convex domain $B$. Stochastic
projected gradient descent on $L$ is
\begin{equation}
\Delta L_{ij}\propto X_{it}X_{jt}-D_{ij}\label{eq:StochasticGradientDescent}
\end{equation}
followed by rectification of $L$. This update preserves symmetry
of $L$, provided that $D$ is symmetric. Gradient ascent-descent
is not generally guaranteed to converge to a steady state, but may
display other dynamical behaviors such as limit cycles. We have found
in practice that convergence to a steady state is not difficult to
obtain, as will be shown by empirical results later on.

\section{Biological interpretation}

We now elaborate on the biological interpretation of the above learning
algorithm, using specific choices for $\Phi\left(W\right)$, convex
domain $B$, and the desired correlation matrix $D$. We define
\begin{equation}
\/\Phi\left(W\right)=\frac{\kappa}{2}\sum_{i}\left(\sum_{a}W_{ia}-\rho\right)^{2}\label{eq:Penalty}
\end{equation}
which is minimized when the row sums of $W$ are equal to $\rho$.
We define $B$ by the bound constraints, $0\leq W_{ia}\leq\omega$.
Since it is online, the algorithm will be rewritten below omitting
time $t$.

Given a stimulus vector $\vec{u},$ update the activities $x_{i}$
of the output neurons using
\begin{align}
x_{i} & :=\frac{1}{L_{ii}}\left[\sum_{a}W_{ia}u_{a}-\sum_{j,j\neq i}L_{ij}x_{j}\right]^{+}\label{eq:DiscreteTimeDynamics}
\end{align}
The dynamics may cycle through the neurons in a fixed or random order.
The input neurons $u_{a}$ try to activate the output neurons $x_{i}$
through the excitatory connections $W_{ia}$. The output neurons try
to turn each other off through the lateral inhibitory connections
$L_{ij}$. Due to the (half-wave) rectification of Eq. (\ref{eq:DiscreteTimeDynamics}),
denoted by $\left[z\right]^{+}=\max\{z,0\}$, neural activities are
nonnegative. This is consistent with the interpretation of $x_{i}$
as neural activity defined by rate of action potential firing (a rate
is nonnegative by definition). 

After convergence of $\vec{x}$, the connection strengths are updated.
The excitatory connections change as

\begin{align}
\Delta W_{ia} & =\eta_{W}\left[x_{i}u_{a}-\kappa\left(\sum_{b}W_{ib}-\rho\right)\right]\label{eq:UpdateW}
\end{align}
The first term is Hebbian, as it causes strengthening of $W_{ia}$
when $x_{i}$ and $u_{a}$ are coactive. The second term weakens $W_{ia}$
when other synapses converging onto the same neuron are strengthened.
It has the effect of creating competition between convergent synapses.
For large $\kappa$, the plasticity rule drives the connections to
satisfy $\sum_{b}W_{ib}\approx\rho$. In words, the connections converging
onto a neuron behave as if they are competing for a fixed resource
$\rho$. Hebbian plasticity determines how strength is allocated between
the connections, but does not change the overall sum of strengths
\citep{miller1996synaptic}. 

After the update (\ref{eq:UpdateW}), the connections are thresholded
to the range $[0,\omega]$ via
\begin{align}
W_{ia} & :=\max\{0,W_{ia}\}\qquad W_{ia}:=\min\left\{ \omega,W_{ia}\right\} \label{eq:RectificationW}
\end{align}
The first thresholding prevents negative connections, preserving the
interpretation of $W$ as excitatory connectivity. The second thresholding
ensures that connections do not exceed the upper bound $\omega$.
As will be shown later, the outcome of competition is simple when
$\kappa$ is large and the ratio $\rho/\omega$ is a positive integer.
Then the number of surviving synapses converging on each neuron is
$\rho/\omega$, and all survivors have maximal strength $\omega$.
The remaining synapses have strength zero, i.e., they have been eliminated.
Similar models of synapse elimination have been used previously in
theoretical studies of neural development \citep{miller1994role}.
This model of synaptic competition is presented because it is especially
simple, but other models produce similar results. For example, Appendix
\ref{sec:AnalogElimination} presents an alternative model in which
the surviving synapses are not fixed in number, and have graded, analog
strengths. Note that there are many other models of synaptic competition
that do not eliminate synapses. Included in this category are models
that hold fixed the Euclidean norm of convergent synapses but lack
nonnegativity constraints \citep{oja1982simplified}. 

The off-diagonal ($i\neq j)$ elements of $L$ are \emph{lateral}
connections, mediating competitive interactions between the output
neurons. They are updated via
\begin{align}
\Delta L_{ij} & =\eta_{L}\left(x_{i}x_{j}-p^{2}\right)\label{eq:UpdateInhibition}
\end{align}
In the first term, coincident activation of presynaptic and postsynaptic
neurons causes strengthening of $L$. This sometimes called ``anti-Hebbian''
plasticity since the $L$ connections are inhibitory \citep{foldiak1989adaptive,rubner1989self,foldiak1990forming,rubner1990development}.
The plasticity rule (\ref{eq:UpdateInhibition}) is symmetric with
respect to interchange of $i$ and $j$. Therefore, $L$ remains a
symmetric matrix for all time, assuming that its initial condition
is symmetric. In the absence of activity, the second term of Eq. (\ref{eq:UpdateInhibition})
causes weakening of lateral connections. The anti-Hebbian update (\ref{eq:UpdateInhibition})
is followed by rectification,

\begin{align}
L_{ij} & :=\max\{L_{ij},0\}\label{eq:RectificationL}
\end{align}
This constrains the matrix $L$ to be nonnegative, so that the lateral
connections remain inhibitory (note the minus sign preceding $L$
in the dynamics of Eq. \ref{eq:DiscreteTimeDynamics}). 

The diagonal elements of $L$ are updated via
\begin{align}
\Delta L_{ii} & =\eta_{L}\left(x_{i}^{2}-q^{2}\right)\label{eq:UpdateHomeostatic}\\
L_{ii} & :=\max\{L_{ii},0\}\label{eq:UpdateHomeostaticLowerBound}
\end{align}
The $L_{ii}$ update seeks to maintain the activity of neuron $i$
at a level set by $q$. Homeostatic regulation of neural activity
has been observed experimentally, and is mediated by more than one
mechanism. The diagonal element $L_{ii}$ is not a synaptic connection
of neuron $i$ to itself (though such connections, known as autapses,
are known to exist). Rather, it has two biological interpretations.
First, it could be regarded as a uniform scaling factor applied to
all synapses, whether feedforward or lateral, that converge on neuron
$i$. Such homeostatic synaptic scaling has been extensively studied
for biological synapses \citep{turrigiano2012homeostatic}. Second,
$L_{ii}$ could be regarded as dividing the activation function in
(\ref{eq:DiscreteTimeDynamics}), acting like a ``gain'' parameter
for neuron $i$. Homeostatic regulation of intrinsic excitability
has also been studied for biological synapses \citep{zhang2003other}. 

After the connection strengths are updated, the dynamics (\ref{eq:DiscreteTimeDynamics})
is run for the next stimulus vector, and so on. The above neural network
learning algorithm is very similar to ones proposed by \citet{foldiak1990forming}
and \citet{pehlevan2014hebbian}. The activity dynamics (\ref{eq:DiscreteTimeDynamics})
uses rectification nonlinearity, as in \citet{pehlevan2014hebbian}.
Appendix \ref{sec:FoldiakOriginal} describes how our formalism can
be extended to the sigmoidal nonlinearity of the \citet{foldiak1990forming}
model. The plasticity rule (\ref{eq:UpdateInhibition}) for the lateral
inhibitory connections is exactly the one in \citet{foldiak1990forming}.
The plasticity rule (\ref{eq:UpdateW}) for the feedforward excitatory
connections includes synaptic competition with elimination. This is
novel relative to \citet{foldiak1990forming} and \citet{pehlevan2014hebbian}
and is drawn from previous models of neural development \citep{miller1994role,miller1996synaptic}. 

In Eq. (\ref{eq:DiscreteTimeDynamics}), the effective synaptic strengths
are given by $W_{ia}/L_{ii}$ and $L_{ij}/L_{ii}$. In both ratios,
the numerator is Hebbian or anti-Hebbian and the denominator is homeostatic.
A similar decomposition of Hebbian and homeostatic plasticity into
multiplicative factors was recently proposed in a model of neural
development \citep{toyoizumi2014modeling}.

\section{Empirical results with MNIST}

\begin{figure}
\begin{centering}
\includegraphics[width=0.8\textwidth]{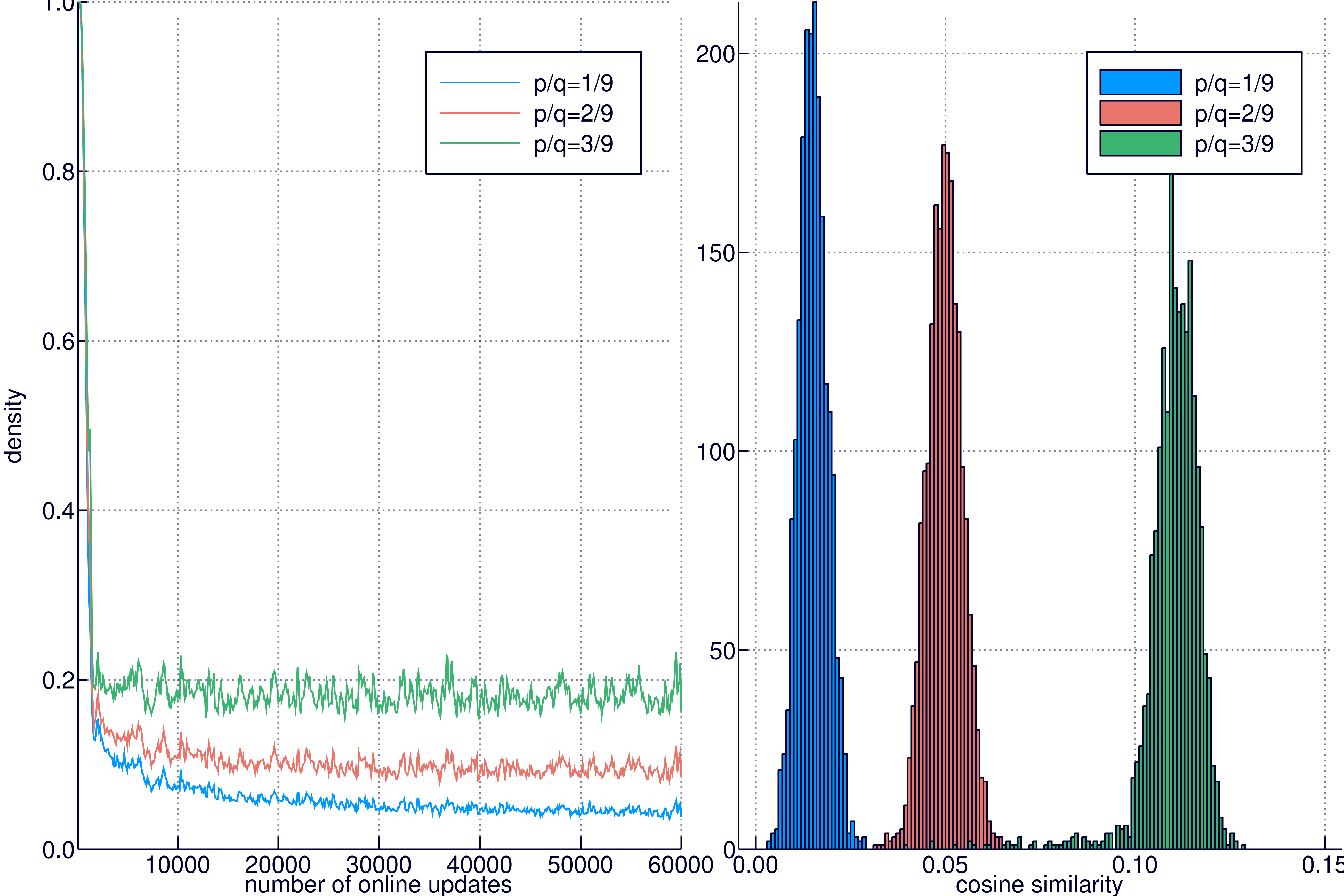}
\par\end{centering}
\caption{Output neuron activity is sparse and decorrelated.\label{fig:SparseActivity}
(a) Fraction of active neurons (activity density) versus time as learning
progresses for 60,000 time steps. The density starts at one, decreases
rapidly at first, and then more slowly. The final value of the density
is lower for smaller $p/q$ . Each point on the curve represents density
averaged over 100 time steps. (b) Cosine similarity of activities
as defined on the left side of Eq. (\ref{eq:CosineSimilarity}). The
vertical coordinate is the number of neuron pairs with that cosine
similarity. Cosine similarity was computed over the last 10,000 time
steps of the learning. In each case, the cosine similarity is clustered
around the value $p^{2}/q^{2}$, as predicted by Eq. (\ref{eq:CosineSimilarity}).
Simulations used $\kappa=1$, $\rho=1$, $\rho/\omega=10$, $q=0.09$,
$\eta_{L}=0.1$, $\eta_{W}=0.001$. In these experiments and all others,
the elements of the $W$ matrix were chosen randomly from a uniform
distribution on $[0,1]$ and then each row was normalized so that
it summed to $\rho$. The matrix $L$ was initialized to the identity
matrix. }
\end{figure}
\begin{figure}
\begin{centering}
\includegraphics[width=0.45\textwidth]{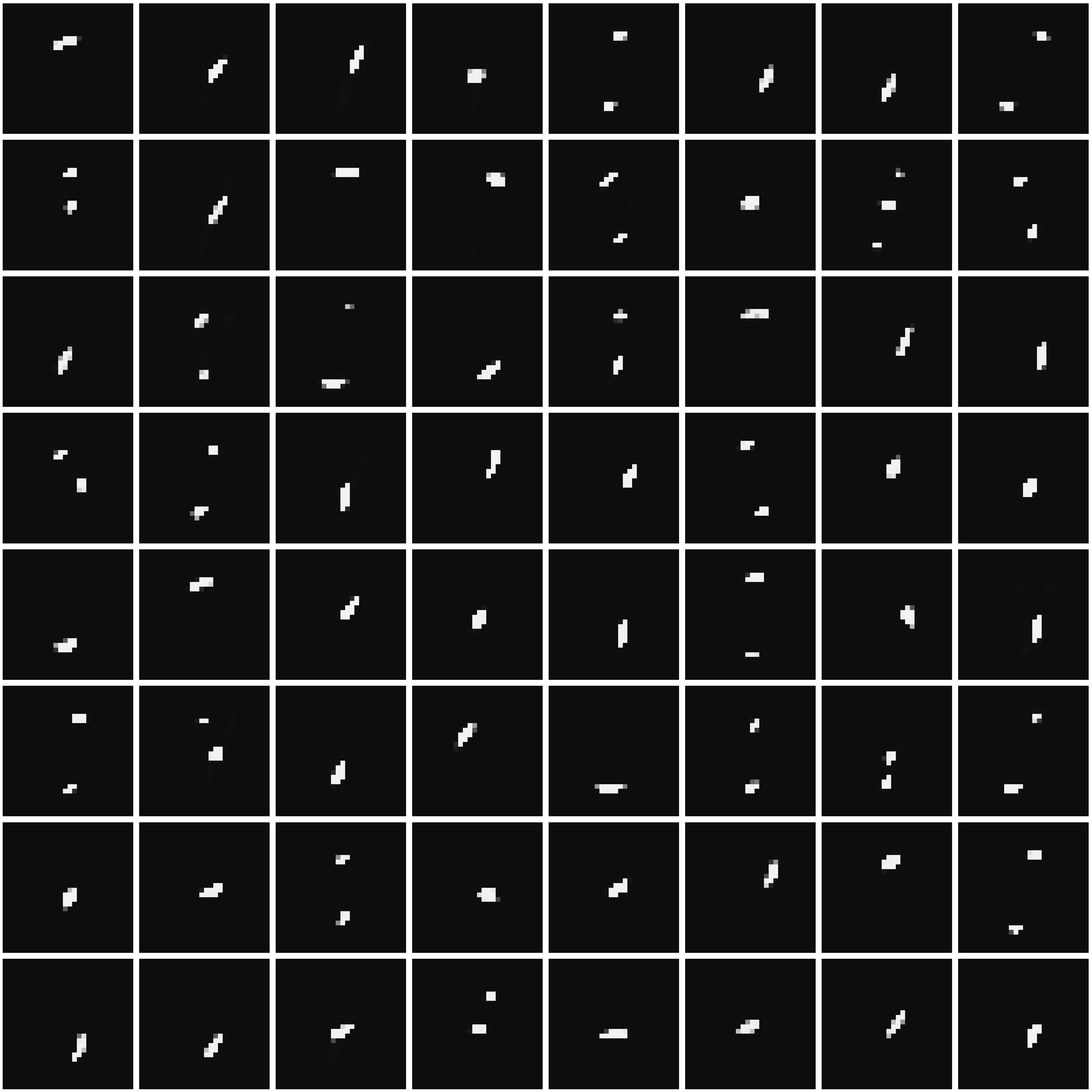}\qquad{}\includegraphics[width=0.45\textwidth]{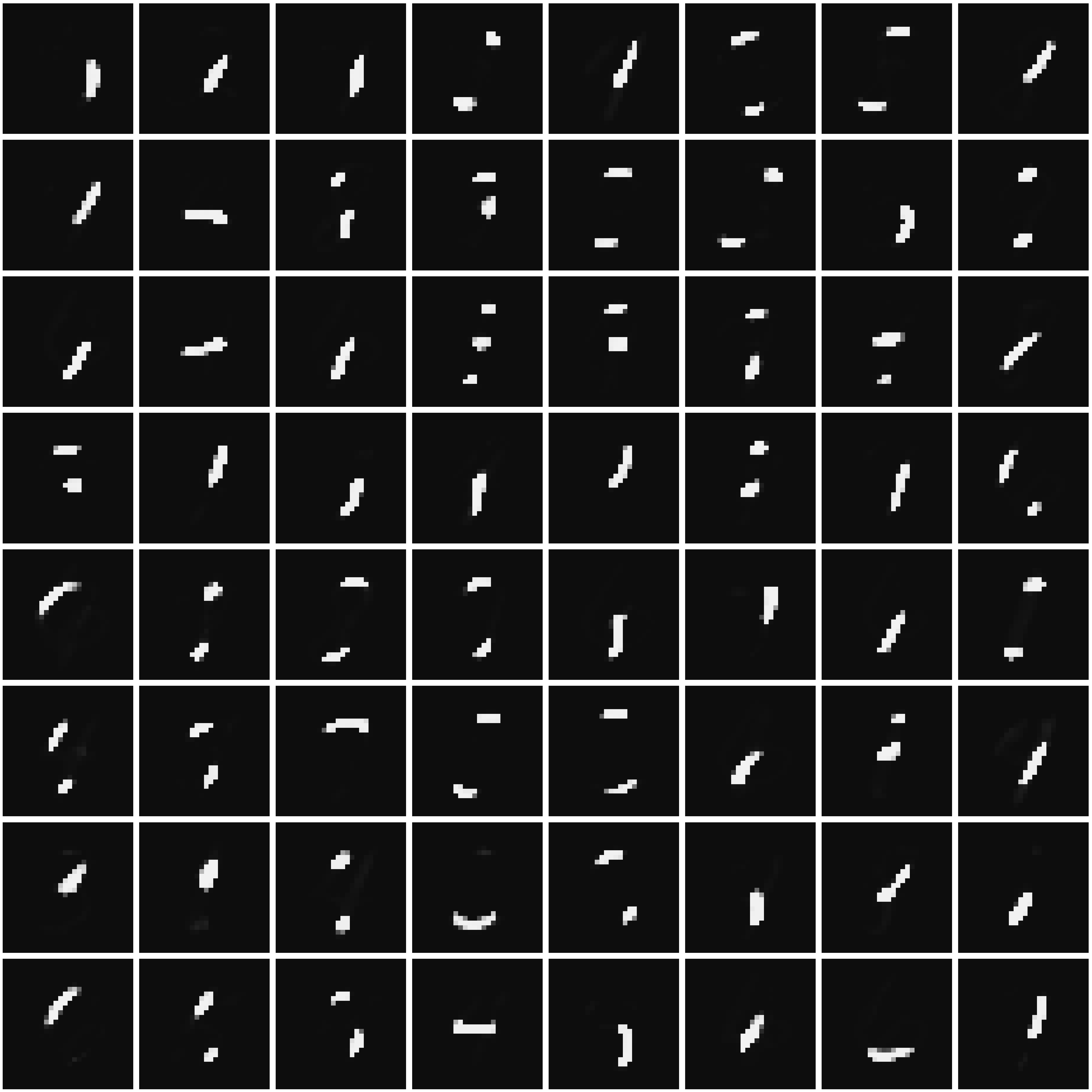}
\par\end{centering}
\caption{Learned connectivity is sparse, as competition between synapses leads
to elimination.\label{fig:SparseConnectivity} The rows of $W$ contain
64 features learned from MNIST and are displayed as images. Experiments
for (a) $\rho/\omega=10$ and (b) $\rho/\omega=20$ demonstrate that
this parameter controls the sparsity of $W$ after learning. Other
parameter settings were $p=0.03$, $q=0.09$, $\kappa=1$ , $\rho=1$,
$\eta_{L}=0.01$, $\eta_{W}=0.001$. Many features consist of spatially
contiguous pixels; this is an outcome of learning as the algorithm
has no prior knowledge of which pixels are near each other.}
\end{figure}
To illustrate the properties of the algorithm, we ran it on the MNIST
dataset. Figure \ref{fig:SparseActivity} shows that sparse activity
patterns emerge from learning. We define density as the fraction of
neurons with nonzero activity. The lower the density, the more sparse
the representation is. The density starts out at 1, and decreases
as learning proceeds. As expected, the final density is smaller when
$p/q$ is smaller. 

The learned connectivity is shown in Figure \ref{fig:SparseConnectivity}.
Most connection strengths have been driven to zero; these synapses
have effectively been eliminated. Each vector of convergent connections
is displayed as an image. The features can be interpreted as ``parts''
of handwritten digits. Increasing $\rho/\omega$ increases the density
of the connectivity. Of course, many other unsupervised learning algorithms
have been used to learn features that resemble parts of objects \citep{lee1997unsupervised,lee1999learning,hoyer2004non}.
The main distinction of the present algorithm is its biological plausibility.
It should be noted that synapse elimination by competition is essential
for sparse connectivity in the present model. Without synaptic competition,
the Hebbian update of Eq. (\ref{eq:UpdateW}) would cause each weight
vector to be a nonnegative superposition of input vectors. The learned
features would resemble whole objects rather than object parts.

\begin{wrapfigure}{o}{0.65\textwidth}%
\includegraphics[width=0.6\textwidth]{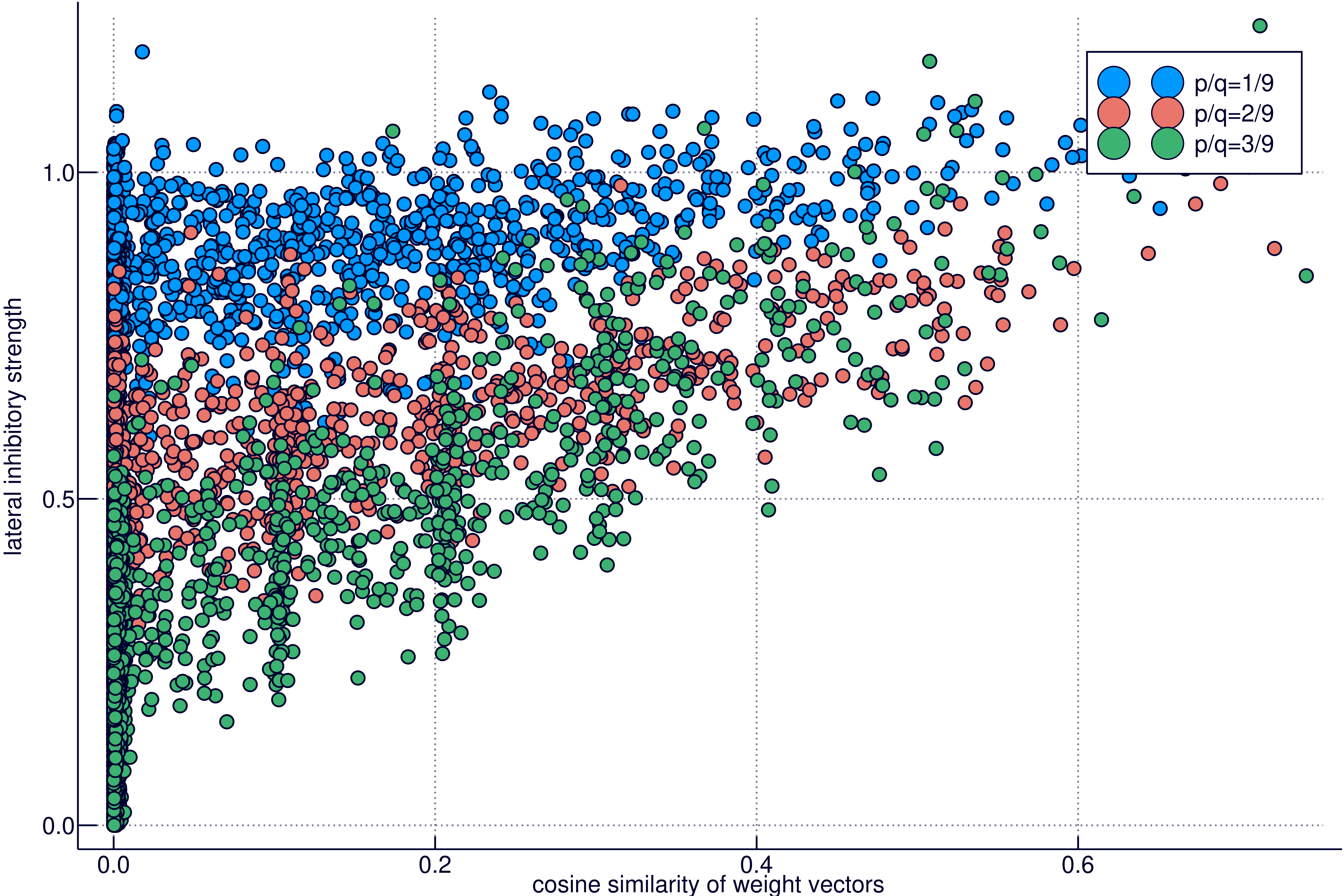}

\caption{Strength of lateral inhibition vs. similarity of weight vectors.\label{fig:InhibitionVsWeightSimilarity}
Each point corresponds to a single pair of neurons. The bunched up
points on the far left represent the many neural pairs that are connected
by strong inhibition yet at the same time have completely different
weight vectors. According to the trend in the other points, neurons
with more similar weight vectors tend to inhibit each other more strongly.}
\end{wrapfigure}%

How are the lateral connections related to feedforward excitation?
If two neurons receive similar feedforward weights, their activities
will be highly correlated in the absence of lateral inhibition. To
weaken the correlation, the learning algorithm is expected to strengthen
the inhibitory connection between the two neurons. Such a trend is
indeed seen empirically in Figure \ref{fig:InhibitionVsWeightSimilarity}.
At the same time, there are many neurons with completely different
weight vectors, but are still connected by strong lateral inhibition.
In the absence of lateral inhibition, the activities of these neurons
would presumably be highly correlated due to the correlations present
in the input. Hence the learning algorithm also strengthens inhibition
to reduce the correlations.

\section{Theoretical analysis of synapse elimination}

Figure \ref{fig:SparseConnectivity} shows that $\rho/\omega$ controls
the number of surviving synapses. For theoretical insight into this
phenomenon, suppose that $\Phi\left(W\right)=\sum_{i}\phi\left(\vec{w}_{i}\right)$
where $\vec{w}_{i}$ is the $i$th column of $W$. Then $\Phi^{\ast}\left(C\right)=\sum_{i}\phi^{\ast}\left(\vec{c}_{i}\right)$
where $\vec{c}_{i}$ is the $i$th column of $C$. For the penalty
function of Eq. (\ref{eq:Penalty}), we have 
\begin{align}
\phi(\vec{w}) & =\frac{\kappa}{2}\left(\sum_{a}w_{a}-\rho\right)^{2}\label{eq:PenaltySingleNeuron}\\
\phi^{\ast}\left(\vec{c}\right) & =\max_{\vec{w}\in B}\left\{ \sum_{a}w_{a}c_{a}-\frac{\kappa}{2}\left(\sum_{a}w_{a}-\rho\right)^{2}\right\} \nonumber 
\end{align}
For simplicity, we consider the limit of infinite $\kappa$, when
$\vec{w}$ lies on the simplex $S^{\rho}$ satisfying $\sum w_{a}=\rho$,
so that 
\[
\phi^{\ast}\left(\vec{c}\right)=\max_{\vec{w}\in B\cap S^{\rho}}\left\{ \sum_{a}w_{a}c_{a}\right\} 
\]
Suppose also that $\rho$ is an integral multiple of $W_{max}$, i.e.,
$\rho=k\omega$. Assume without loss of generality that $c_{1}\geq c_{2}\geq\ldots\geq c_{N}$.
Then the maximum is achieved by setting $w_{1}=\cdots=w_{k}=\omega$
and $w_{k+1}=\cdots=w_{N}=0$, and
\[
\phi^{\ast}\left(\vec{c}\right)=\omega\sum_{a=1}^{k}c_{a}
\]
In other words, the objective function $\Phi^{\ast}$ in Eqs. (\ref{eq:Primal})
and (\ref{eq:Dual}) is proportional to the sum of the top $k$ elements
of each row of the output-input correlation matrix.

Recall that for a single neuron, the correlation game reduces in Eq.
(\ref{eq:MaximumVariance}) to finding a projection with maximal second
moment, subject to the penalty function $\phi$. For the specific
case of the penalty function above, this is a version of sparse PCA
\citep{moghaddam2006spectral,zass2007nonnegative} in which the projection
vector $\vec{w}$ is required to have $k$ elements equal to one and
all other elements equal to zero. Effectively, this selects the set
of $k$ inputs that have maximal correlations with each other.

\section{Discussion}

Our correlation game is a new formulation of unsupervised learning
that is convenient for understanding biologically plausible synaptic
plasticity rules in neural networks. We have considered several intuitions
about the computational functions of synaptic plasticity:
\begin{enumerate}
\item Hebbian feedforward excitation enables neurons to learn features from
sensory inputs.
\item Anti-Hebbian inhibition serves to sparsen activity, facilitating the
learning of features by Hebbian excitation.
\item By sparsening connectivity, synaptic competition and elimination facilitate
the learning of features that resemble parts of objects.
\end{enumerate}
These long-standing intuitions have been made mathematically precise
using the correlation game, and are illustrated by the numerical experiments
of Figs. \ref{fig:SparseActivity} and \ref{fig:SparseConnectivity}.

The neural network learns by fostering competition (1) between neurons
mediated by lateral inhibition, (2) between synapses converging onto
the same neuron, and (3) between excitation and inhibition in the
correlation game.

\section*{Acknowledgments}

We are grateful to Mitya Chklovskii and Cengiz Pehlevan for illuminating
discussions on game theoretic formalisms for neural networks with
Hebbian and anti-Hebbian plasticity. We thank Ken Miller for clarifications
concerning models of neural development.

\bibliographystyle{plainnat}
\bibliography{Foldiak}

\appendix

\section{Relations with other unsupervised learning principles\label{sec:OtherObjectives}}

Unsupervised learning has been formulated as maximization of the mutual
information between output and input \citep{linsker1988self,atick1990towards,bell1995information},
or minimization of the mutual information between outputs \citep{hyvarinen2000independent}.
Our output-input correlations can be viewed as a proxy for mutual
information between output and input. Our constraint keeps the output-output
correlations small, which is similar to limiting the redundancy of
the outputs, or the mutual information between the outputs. We prefer
correlations rather than entropies in our unsupervised learning principle,
because we would like to derive Hebbian plasticity rules. There is
no explicit model of the relationship between input and output in
Eq. (\ref{eq:ConstrainedOptimization}); the relationship will emerge
from solving the optimization problem. In contrast, most information
theoretic formulations explicitly assume a linear relationship between
output and input.

Unsupervised learning has been formulated as projection onto a subspace
that maximizes a moment-based statistic like variance \citep{oja1982simplified,linsker1988self}
or kurtosis \citep{huber1985projection}. Our Problem \ref{eq:ConstrainedOptimization}
is equivalent for the case of a single output, $X_{it}=x_{t}$. To
demonstrate this, combine Eqs. (\ref{eq:ConstrainedOptimization})
and (\ref{eq:ConvexConjugate}) to yield
\begin{equation}
\max_{\vec{w}\in B}\max_{X\geq0}\left\{ \frac{1}{T}\sum_{t}x_{t}\sum_{a}w_{a}U_{at}-\phi\left(\vec{w}\right)\right\} \text{ such that }\frac{1}{T}\sum_{t}x_{t}^{2}\leq q^{2}\label{eq:SingleNeuron}
\end{equation}
(The notation has switched to $\phi$ from $\Phi$ to indicate a function
of a vector $\vec{w}$ rather than a matrix $W$, and the scalar $q^{2}$
denotes the single element of the $1\times1$ matrix $D$.) The maximum
with respect to $X$ is $x_{t}\propto\sum_{a}w_{a}U_{at},$ i.e.,
the output is proportional to a projection of the input along the
direction defined by $\vec{w}$. Then the remaining optimization depends
on the input through the second moment $\langle\left(\vec{w}\cdot\vec{u}\right)^{2}\rangle$.
\begin{equation}
\max_{\vec{w}\in B}\left\{ q\sqrt{\frac{1}{T}\sum_{t}\left(\sum_{a}w_{a}U_{at}\right)^{2}}-\phi\left(\vec{w}\right)\right\} =\max_{\vec{w}\in B}\left\{ q\sqrt{\langle\left(\vec{w}\cdot\vec{u}\right)^{2}\rangle}-\phi\left(\vec{w}\right)\right\} \label{eq:MaximumVariance}
\end{equation}
If we choose $\phi(\vec{w})=0$ and $B$ the unit ball, this amounts
to finding a one-dimensional projection with maximal second moment.
For zero-centered inputs, this is equivalent to maximizing variance,
which yields the principal component. 

It is well-known that a single linear neuron with Hebbian synapses
learns the principal component of the input distribution \citep{oja1982simplified}.
A problem with generalizing this idea to more than one principal component
is that $k$ linear neurons have the tendency to all learn the same
principal component rather than the top $k$ distinct principal components
\citep{sanger1989optimal}. Preventing such similarity could be regarded
as the motivation for Eq. (\ref{eq:CosineSimilarity}) with $p/q<1$,
and was the original intuition behind the model of \citet{foldiak1990forming}.
In the general case of multiple outputs, the relationship between
$X$ and $U$ is nonlinear, and will be derived below.

Finally, our Problem \ref{prob:Primal} can be viewed as a generalization
of the similarity matching principle of \citet{pehlevan2015hebbian}.
In classical multidimensional scaling (CMDS), the objective is to
minimize the sum of squared differences between output and input similarities
at all pairs of times, 
\begin{equation}
\frac{1}{2}\sum_{tt'}\left[\vec{x}(t)\cdot\vec{x}(t')-\vec{u}(t)\cdot\vec{u}(t')\right]^{2}=\frac{1}{2}\left\Vert X^{\top}X-U^{\top}U\right\Vert ^{2}\label{eq:CMDSCost}
\end{equation}
When $\vec{x}$ has lower dimensionality than $\vec{u},$ the CMDS
cost function is minimized by $X$ that is a projection of $U$ onto
the subspace spanned by the principal components of the input \citep{williams2002connection}.
\citet{pehlevan2015hebbian} used this idea to derive a neural network
learning algorithm for implementing PCA. In the special case of $\Phi\left(W\right)=\sum_{ia}W_{ia}^{2}/2$
and $B$ the nonnegative orthant, the maximization of Eq. (\ref{eq:ConvexConjugate})
yields $\Phi^{\ast}\left(C\right)=\sum_{ia}C_{ia}^{2}/2,$ since $C$
is assumed nonnegative. Given the constraint $XX^{\top}/T\leq D$,
the CMDS cost function can be upper bounded by
\begin{align*}
\frac{1}{2}\left\Vert X^{\top}X-U^{\top}U\right\Vert ^{2} & =\frac{1}{2}\left\Vert XX^{\top}\right\Vert ^{2}-\left\Vert XU^{\top}\right\Vert ^{2}+\frac{1}{2}\left\Vert UU^{\top}\right\Vert ^{2}\\
 & \leq\frac{T^{2}}{2}\left\Vert D\right\Vert ^{2}-\left\Vert XU^{\top}\right\Vert ^{2}+\frac{1}{2}\left\Vert UU^{\top}\right\Vert ^{2}
\end{align*}
Therefore Eq. (\ref{eq:ConstrainedOptimization}) can be viewed as
the minimization of an upper bound for the CMDS cost function (\ref{eq:CMDSCost})
given the constraint $XX^{\top}/T\leq D$.

\section{Extension to sigmoidal nonlinearity\label{sec:FoldiakOriginal}}

\citet{foldiak1990forming} used sigmoidal rather than rectification
nonlinearity in his neural network model. Our formalism can be extended
to sigmoidal nonlinearity as follows. Problem \ref{prob:Primal} is
replaced by

\[
\max_{X}\Phi^{\ast}\left(\frac{1}{T}XU^{T}\right)-\frac{1}{T}\sum_{it}\bar{F}\left(X_{it}\right)
\]
such that 
\[
\frac{1}{T}\sum_{t}X_{it}=p\text{ for all }i
\]
\[
\frac{1}{T}\sum_{t}X_{it}X_{jt}\leq p^{2}\text{ for all }i\neq j
\]
Then Lagrangian duality yields the analog of Problem \ref{prob:Dual},
\[
\max_{X}\min_{L\geq0}\min_{\theta}\left\{ \Phi^{\ast}\left(\frac{1}{T}XU^{T}\right)-\frac{1}{T}\sum_{it}\bar{F}\left(X_{it}\right)-\theta_{i}\left(\frac{1}{T}\sum_{t}X_{it}-p\right)-\sum_{i<j}L_{ij}\left(\frac{1}{T}\sum_{t}X_{it}X_{jt}-p^{2}\right)\right\} 
\]
or equivalently

\[
\frac{1}{T}\max_{X}\min_{L\geq0}\min_{\theta}\left\{ \sum_{t}\left[\sum_{ia}W_{ia}X_{it}U_{at}-\Phi\left(W\right)-\sum_{i}\bar{F}\left(X_{it}\right)-\theta_{i}\left(X_{it}-p\right)-\sum_{i<j}L_{ij}\left(X_{it}X_{jt}-p^{2}\right)\right]\right\} 
\]
The coordinate ascent dynamics (\ref{eq:CoordinateAscent}) is replaced
by
\begin{equation}
X_{it}:=f\left(\sum_{\alpha}W_{ia}U_{at}-\sum_{j,j\neq i}L_{ij}X_{jt}-\theta_{i}\right)\label{eq:DynamicsSigmoidal}
\end{equation}
where $\overline{F}'=f^{-1}$. The updates (\ref{eq:StochasticGradientAscent})
and (\ref{eq:StochasticGradientDescent}) for $W$ and the off-diagonal
elements of $L$ remain the same. However, the update rule for $L_{ii}$
is replaced by an update for the threshold $\theta_{i}$ in Eq. (\ref{eq:DynamicsSigmoidal}),

\begin{equation}
\Delta\theta_{i}\propto X_{it}-p\label{eq:ThresholdUpdate}
\end{equation}
which can be viewed as an alternate model for homeostatic regulation
of activity. 

In the \citet{foldiak1990forming} model, when combined with homeostatic
plasticity as in Eq. (\ref{eq:ThresholdUpdate}), the anti-Hebbian
update of Eq. (\ref{eq:UpdateInhibition}) acts to drive the covariances
of all pairs of neural activities to zero. In other words, the neural
activities are approximately statistically independent. Consistent
with this idea, \citet{falconbridge2006simple} argued that the \citet{foldiak1990forming}
algorithm implements a kind of independent component analysis. Our
algorithm makes the cosine similarities small when $p/q$ is small.
Therefore one might call it ``desimilarizing'' rather than decorrelating.
Another difference is that we impose nonnegativity constraints on
both feedforward and lateral connections, while \citet{foldiak1990forming}
only imposed constraints on the latter.

\section{Relation to contrastive Hebbian learning}

Of the existing neural net learning algorithms, the one that hews
most closely to Hebbian principles is the Boltzmann machine or contrastive
Hebbian learning. The inhibitory plasticity of the present model is
actually a form of contrastive Hebbian learning. This can be seen
by defining 

\[
E\left(X\right)=\frac{1}{T}\sum_{t}\left(\frac{1}{2}\sum_{ij}L_{ij}X_{it}X_{jt}-\sum_{ia}W_{ia}X_{it}U_{at}\right)
\]
and rewriting Eq. (\ref{eq:Payoff}) as

\begin{equation}
R\left(W,L\right)=\left[E\left(X^{0}\right)-\min_{X\geq0}E(X)\right]+\frac{1}{T}\sum_{ia}\left[W_{ia}X_{it}^{0}U_{at}-\Phi\left(W\right)\right]\label{eq:ObjectiveFunctionContrastive}
\end{equation}
where the ``ghost activity'' matrix $X^{0}$ satisfies $X^{0}X^{0\top}/T=D$.
The gradient update for $L$ involves the first term only, since the
second term does not depend on $L$. The first term of Eq. (\ref{eq:ObjectiveFunctionContrastive})
is the cost function for contrastive Hebbian learning. The columns
of the ghost activity matrix represent the ``clamped'' or desired
patterns for the learning, while the vectors $X$ represent the ``free''
patterns. 

The preceding makes clear that the plasticity rule for the inhibitory
connections is the same as in contrastive Hebbian learning \citep{movellan1991contrastive,baldi1991contrastive},
except here the nonlinearity is rectification rather than sigmoidal.
Plasticity of inhibition is trying to make the second order statistics
of the neural activities match those of the ghost activities. 

\section{Synapse elimination without upper bounds\label{sec:AnalogElimination}}

\begin{figure}
\includegraphics[width=0.9\textwidth]{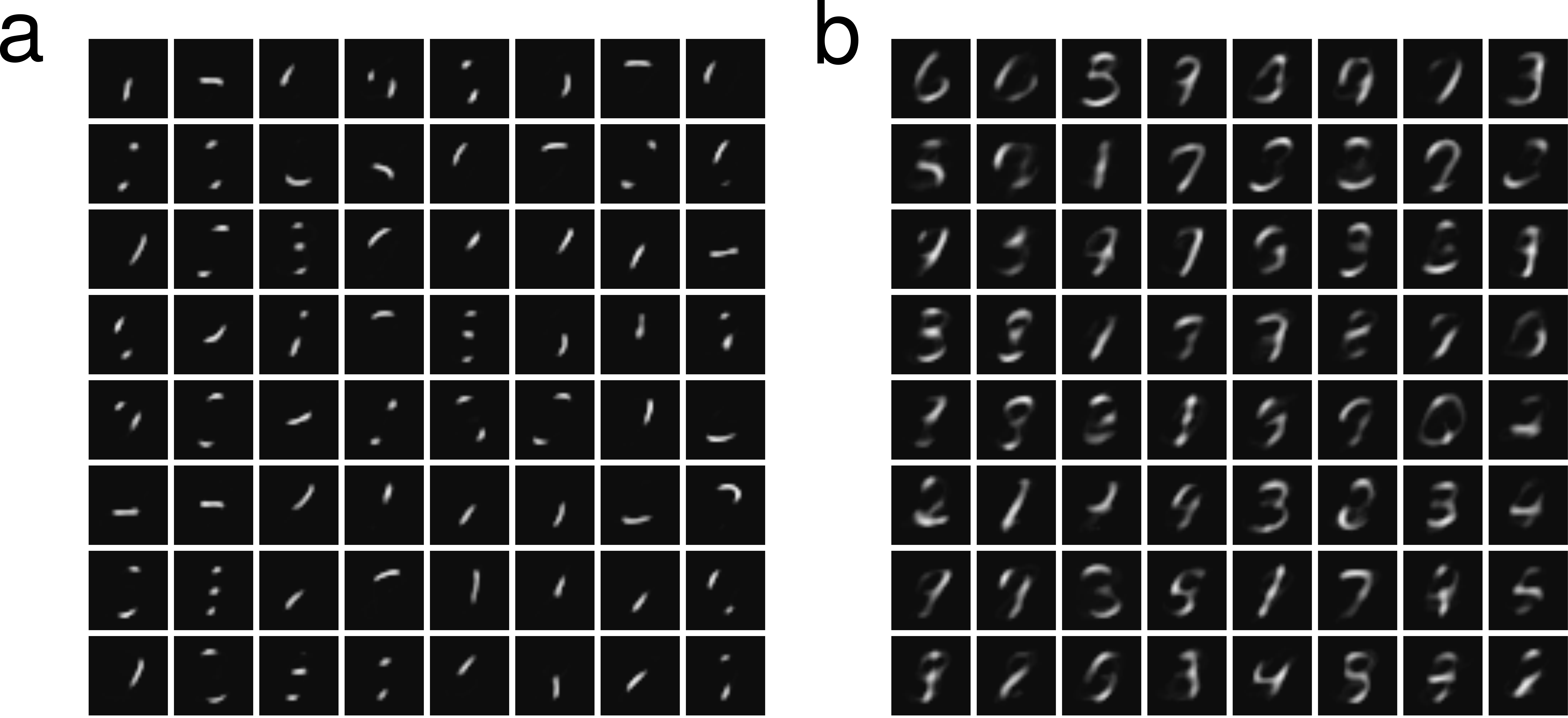}

\caption{Synaptic competition with elimination still occurs after replacing
the upper bound on synaptic strength (Eq. \ref{eq:UpdateW}) with
a simple weight decay (Eq. \ref{eq:UpdateWAnalogElimination}).\label{fig:SparseConnectivityAnalog}
The rows of $W$ contain 64 features learned from MNIST and are displayed
as images. Experiments for (a) $\gamma=0.1$ and (b) $\gamma=0.5$
demonstrate that this parameter controls the sparsity of $W$ after
learning. The results are qualitatively similar to those of Fig. \ref{fig:SparseConnectivity},
except that synaptic weights are graded rather than saturated at upper
and lower bounds. Other parameter settings were $p=0.03$, $q=0.09$,
$\kappa=1$, $\rho=1$, $\eta_{L}=0.1$, $\eta_{W}=0.001$.}
\end{figure}

Eq. (\ref{eq:RectificationW}) contains an upper bound $\omega$ for
all excitatory synaptic strengths. The number of surviving synapses
is set by $\rho/\omega$. If we replace Eq. (\ref{eq:UpdateW}) by
a more complex model of synaptic competition

\begin{align}
\Delta W_{i\alpha} & \propto x_{i}u_{\alpha}-\gamma W_{i\alpha}-\kappa\left(\sum_{\beta}W_{i\beta}-\rho\right)\label{eq:UpdateWAnalogElimination}
\end{align}
then synapse elimination is possible without the upper bound $\omega$
(though there is still an upper bound on summed synaptic strength
due to $\rho$). As can be seen from Fig. \ref{fig:SparseConnectivityAnalog},
the results of learning are qualitatively similar. The parameter $\gamma$
controls the sparsity of $W$, with smaller values producing ``parts''
and larger values producing ``wholes.'' The synaptic weights are
graded, rather than saturated at upper and lower bounds as in Fig.
\ref{fig:SparseConnectivity}. 

A theoretical analysis of this analog model of synaptic competition
is more complex, but the end result is qualitatively similar to the
previous bound-constrained model. As we will see below, the competition
eliminates connections from weakly correlated inputs, focusing the
objective function on strongly correlated inputs. The $W$ update
of Eq. (\ref{eq:UpdateWAnalogElimination}) can be derived from the
penalty function

\begin{equation}
\Phi\left(W\right)=\frac{\gamma}{2}\sum_{ia}W_{ia}^{2}+\frac{\kappa}{2}\sum_{i}\left(\sum_{a}W_{ia}-\rho\right)^{2}\label{eq:PenaltyL2}
\end{equation}
The convex conjugate can be written as 
\begin{align}
\overline{\phi}\left(\vec{c}\right) & =\max_{\vec{w}\geq0}\left\{ \sum_{a}w_{a}c_{a}-\frac{\gamma}{2}\sum_{a}w_{a}^{2}-\frac{\kappa}{2}\left(\sum_{a}w_{a}-\rho\right)^{2}\right\} \label{eq:ObjectiveVector}
\end{align}
summed over the rows of the matrix $C$. According to the KKT conditions
for the optimum, either
\begin{equation}
w_{a}\geq0\text{ and }c_{a}-\gamma w_{a}-\kappa\left(\sum_{b}w_{b}-\rho\right)=0\label{eq:Survive}
\end{equation}
or
\begin{equation}
w_{a}=0\text{ and }c_{a}-\gamma w_{a}-\kappa\left(\sum_{b}w_{b}-\rho\right)\leq0\label{eq:Eliminate}
\end{equation}
must be satisfied for all $a$. The conditions can be combined into
an expression of the form $\gamma w_{a}=\left[c_{a}-\theta\right]^{+}$
where the threshold is given by $\theta=\kappa\left(\sum_{b}w_{b}-\rho\right)$.
Assume without loss of generality that $c_{1}\geq c_{2}\geq\ldots\geq c_{N}$.
Then the solution of the KKT conditions is
\[
\gamma w_{a}=\left[c_{a}-\theta_{k}\right]^{+}
\]
where
\[
\theta_{k}=\kappa\left(\sum_{a=1}^{k}w_{a}-\rho\right)=\frac{1}{k+\gamma/\kappa}\left(\sum_{a=1}^{k}c_{a}-\rho\gamma\right)
\]
satisfies the inequality
\begin{equation}
c_{k}>\theta_{k}\geq c_{k+1}\label{eq:SynapseSurvivalNumber}
\end{equation}
At the solution, $w_{1},\ldots,w_{k}$ are positive while $w_{k+1},\ldots,w_{N}$
vanish. In the context of the learning algorithm, this means that
$k$ synapses survive while the rest are eliminated. The surviving
synapses come from the inputs that are most correlated with the output.

In the limit of infinite $\kappa$, Eq. (\ref{eq:SynapseSurvivalNumber})
yields some simple conditions for synapse elimination. At least one
synapse is eliminated if and only if

\[
\rho\gamma<\sum_{b=1}^{N}c_{b}-Nc_{N}=N(\bar{c}-c_{N})
\]
In other words, synapse elimination is triggered if and only if the
quotient $\rho\gamma/N$ is smaller than the difference between the
mean $\bar{c}$ and the minimum $c_{N}$ of the elements of $\vec{c}$.
It can also be shown that the solution of the inequality (\ref{eq:SynapseSurvivalNumber})
is $k=1$ for sufficiently small $\rho\gamma$. This is winner-take-all
competition: all synapses are eliminated except for a single winner. 

What is the value of the objective function $\overline{\phi}$ at
the optimum? If $\kappa=0$, the objective function reduces to
\begin{equation}
2\gamma\overline{\phi}\left(\vec{c}\right)=\left\Vert \vec{c}\right\Vert ^{2}=\sum_{a}c_{a}^{2}\label{eq:Frobenius}
\end{equation}
In this case, learning is trying to maximize the sum of the squares
of the output-input correlations. This is the objective function used
by the similarity matching principle of \citet{pehlevan2015hebbian}. 

For $\kappa>0$, the result of the optimization is
\[
2\gamma\overline{\phi}\left(\vec{c}\right)=\sum_{a=1}^{k}\left(c_{a}^{2}-\theta_{k}^{2}\right)-\frac{\gamma}{\kappa}\theta_{k}^{2}
\]
where $k$ is the number of surviving synapses. Now the sum only includes
the top correlations, those corresponding to surviving synapses. This
can also be written as $2\gamma\overline{\phi}\left(\vec{c}\right)=\sum_{a}\left[c_{a}^{2}-\theta_{k}^{2}\right]^{+}-\gamma\theta_{k}^{2}$,
so effectively the correlations are being thresholded.

\end{document}